\documentclass[10pt,twocolumn,letterpaper]{article}

\usepackage{wacv}
\usepackage{times}
\usepackage{epsfig}
\usepackage{graphicx}
\usepackage{amsmath}
\usepackage{amssymb}
\usepackage{booktabs}

\usepackage{comment}
\usepackage{tikz}
\usepackage{caption}
\usepackage{subcaption}
\usepackage{booktabs}
\usepackage{pifont}
\newcommand{\cmark}{\ding{51}}%
\newcommand{\norm}[1]{\left\lVert#1\right\rVert}

\newcommand\blfootnote[1]{%
  \begingroup
  \renewcommand\thefootnote{}\footnote{#1}%
  \addtocounter{footnote}{-1}%
  \endgroup
}
\def\wacvPaperID{291} 

\wacvapplicationstrack 

\wacvfinalcopy 

\ifwacvfinal
\usepackage[breaklinks=true,bookmarks=false]{hyperref}
\else
\usepackage[pagebackref=true,breaklinks=true,colorlinks,bookmarks=false]{hyperref}
\fi

\begin{document}

\title{DRAMA: Joint Risk Localization and Captioning in Driving}

\author{
Srikanth Malla$^{1*}$
\and
Chiho Choi$^{1**}$
\and
Isht Dwivedi$\dagger$
\and
Joon Hee Choi$\ddagger$ 
\and
Jiachen Li $\ddagger\ddagger$
\\[-3.0ex]
}

\maketitle
\thispagestyle{empty}

\begin{abstract}
Considering the functionality of situational awareness in safety-critical automation systems, the perception of risk in driving scenes and its explainability is of particular importance for autonomous and cooperative driving. Toward this goal, this paper proposes a new research direction of joint risk localization in driving scenes and its risk explanation as a natural language description. Due to the lack of standard benchmarks, we collected a large-scale dataset, DRAMA (Driving Risk Assessment Mechanism with A captioning module), which consists of 17,785 interactive driving scenarios collected in Tokyo, Japan. Our DRAMA dataset accommodates video- and object-level questions on driving risks with associated important objects to achieve the goal of visual captioning as a free-form language description utilizing closed and open-ended responses for multi-level questions, which can be used to evaluate a range of visual captioning capabilities in driving scenarios. We make this data available to the community for further research. Using DRAMA, we explore multiple facets of joint risk localization and captioning in interactive driving scenarios. In particular, we benchmark various multi-task prediction architectures and provide a detailed analysis of joint risk localization and risk captioning. The data set is available at \url{https://usa.honda-ri.com/drama} \blfootnote{*{\tt\small srikanth@kinetic.auto}, Kinetic Automation\\
\hspace*{1.7em}$**${\tt\small chihochoi@outlook.com}, Samsung Semiconductor\\
\hspace*{1.7em}$\dagger${\tt\small idwivedi@honda-ri.com}, Honda Research Institute\\
\hspace*{1.7em}$\ddagger${\tt\small jh4.choi@samsung.com}, Samsung Semiconductor\\
\hspace*{1.7em}$\ddagger\ddagger${\tt\small jiachen\_li@stanford.edu}, Stanford University\\
\hspace*{1.7em}1 the work was majorly done at Honda Research Institute, USA}
\end{abstract}
\vspace{-0.3cm}
\section{Introduction}
Situational awareness is an important requirement to achieve high level automation in intelligent vehicles. An important aspect of situational awareness for intelligent mobility is the ability to create an explainable network for the perception of risk from the view-point of the driver, and to establish methods to communicate those risks through voice commands for cooperative and effective driving~\cite{keyes2019priming}.  In this sense, conveying what the network understood as perceived risk through caption generation can serve two functions: an explainable model for identifying risk agents in the scene, as well as the ability to generate linguistic descriptions of the scene to communicate those risks through an AI agent.  

Important object identification is becoming a key ingredient in autonomous driving and driving assistant systems for safety-critical tasks such as behavior prediction, decision making, and motion planning. One set of approaches have been training the model in a self-supervised manner in addition to their original tasks (e.g., trajectory prediction or steering control), while other approaches have been directly training their model using the human annotated object-level importance. Although explainability of the network predictions is of particular importance, none of existing methods characterizes captioning about identified important objects in the form of a natural language description. Addressing this need, our problem formulation aims to jointly supervise the explainability that is consistent with identification of important objects.

In the absence of an appropriate dataset for this task, we introduce a new dataset to the community to facilitate the study of producing free-form captions explaining the reasoning behind the designation of important objects. A natural language description interprets the interaction of an important object with the ego-vehicle and reasons about its risk observed from ego-driver's perspective in driving scenarios. Our setting of interest and our new dataset, named Driving Risk Assessment Mechanism with A captioning module (DRAMA), are captured through question and answering, seeking closed and open-ended video- and object-level attributes from sequence and motion observations. 

This dataset provides a basis for a comprehensive exploration of (i) identifying important objects that may influence the future behavior of ego-vehicle; and (ii) captioning about why the system holds its beliefs about future risks of these objects. We designed DRAMA with an explicit goal of visual captioning, particularly in driving scenarios. Each video clip depicts a ego-driver's behavioral response to perceived risk that activates braking of the vehicle. We provide a detailed analysis of such events through questions that are made by elementary operations (what, which, where, why, how) and through answers that represent video- and object-level attributes with a closed and open-ended response, which are utilized to create a free-form caption. Our design choices would encourage in-depth understanding of risks in driving scenarios and allow us to focus on the captioning of these events. To this end, we present algorithmic baselines to characterize the  captioning of perceived risks associated with important agents in various multi-tasking settings. The main contributions of the paper are summarized as follows:

\begin{itemize}
    \item We introduce a novel dataset that provides linguistic descriptions (with the focus on reasons) of driving risks associated with important objects and that can be used to evaluate a range of visual captioning capabilities in driving scenarios. 
    \item We present a data annotation strategy that accommodates the unique characteristics of video- and object-level attributes obtained by closed and open-ended responses.
    \item We provide a new research direction of joint risk localization in driving scenes and its caption as a natural language description and benchmark several baseline models on our DRAMA dataset for these tasks.
\end{itemize} 

\section{Related Work}

\noindent
\textbf{Important Object Identification}\;\; 
Important agent identification methods can be broadly classified as explicit learning and implicit learning.
Explicit learning methods like \cite{zeng2017agent,gao2019goal} formulate the agent importance estimation as a binary classification problem. These types of models are typically trained by standard supervised learning. 
The approaches proposed in \cite{alletto2016dr,tawari2018learning,xia2018predicting} learn to imitate human gaze behavior and predict a pixel-level attention map which is used as a proxy for risk.
Implicit learning methods proposed in \cite{malla2020titan,kim2017interpretable,wang2019deep,li2020make} are trained to perform a related task such as trajectory prediction or steering control. 
Intermediate network activation values are interpreted as perceived risk in these cases.
More specifically, the model in \cite{malla2020titan} is trained to perform future trajectory prediction using self-attention on all the agents in the scene. 
Self-attention activation outputs are used as agent importance estimates. Causal intervention approach proposed in \cite{li2020make} assumes the counter-examples have fixed ground truth. The approaches in \cite{kim2017interpretable,wang2019deep} are trained to predict steering control and pixel-level attention heat maps which are used to estimate agent importance. However, none of these methods addresses the reasoning about the model decision nor provides description in the form of natural language, which hinders humans' understanding and the interpretability of safety-critical autonomous driving (AD) and advanced driver-assistance systems (ADAS).

\noindent
\textbf{Vision and Language Grounding}\;\; 
There exist two major categories in vision and language grounding: object retrieval from captions and dense image captioning. In the former, the algorithms such as \cite{deruyttere2019talk2car,hu2016natural} detect a bounding box in the image given the caption description. Such models for grounding of the object using natural language have been trained with either a fully supervised~\cite{rohrbach2016grounding,chen2017query} or weakly supervised~\cite{zhao2018weakly} manner using Flickr30K entities dataset~\cite{plummer2015flickr30k}. 
In the latter, dense image captioning methods \cite{johnson2016densecap,yin2019context,yang2017dense} find several region proposals and predict the caption associated with each proposal using the Visual Genome dataset~\cite{krishna2017visual}. Some other works apply visual attention to the image for the task of caption generation~\cite{xu2015show} or end-to-end learning of driving actions ~\cite{kim2019grounding}. 
Our problem setting is closely related to dense image captioning as the language description is generated while detecting the corresponding object.  
However, our scope is specifically focused on localizing risks in driving scenes where captions provide the reason behind identified important objects. 

\noindent
\textbf{Datasets}\;\; 
There exist multiple datasets for accident localization~\cite{chan2016anticipating,zeng2017agent,fang2019dada} or anomaly detection~\cite{yao2020and} or causal object localization~\cite{ramanishka2018toward} in driving scenes. Although these benchmarks can be used to understand and predict accidents and related risk attributes, they are inherently incapable of explaining the reason behind important object identification. In-depth analyses of visual explanations can be acquired with the help of linguistic descriptions that explain the perceived risk through a free-form caption. In this view, our DRAMA dataset has unique functions of explanations in the form of a natural language description that is associated with risk localization. 

\begin{figure*}[t!]
    \includegraphics[width=\textwidth]{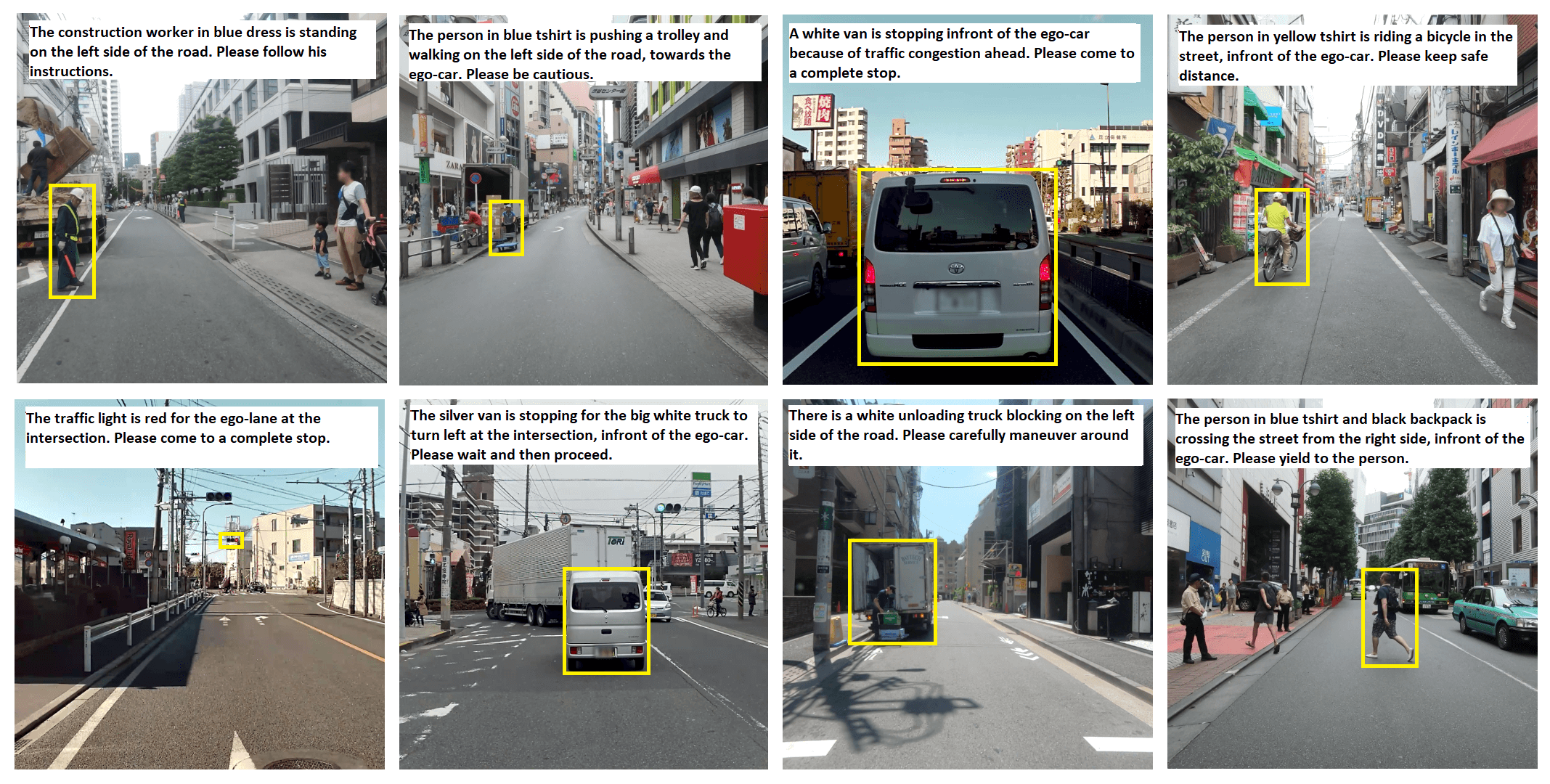}
    \caption{Example Scenes from DRAMA Dataset, more examples including VQA are available in the supplementary.}
    \label{fig:drama_data}
    \vspace{-0.4cm}
\end{figure*}

Few driving datasets are aimed to provide commands written in natural language for autonomous vehicles. Talk2Car~\cite{deruyttere2019talk2car} adds free-form captions on top of the nuScenes dataset \cite{caesar2020nuscenes} to guide the future path of the autonomous car using referred objects in the scene. BDD-X~\cite{kim2018textual} and BDD-OIA~\cite{xu2020explainable} has free-form and closed-form explanations respectively, as reasons for ego-vehicle actions. Similarly, HAD~\cite{kim2019grounding} provides the vehicle with natural language commands to generate salient maps learned from driver's gaze information. However, these driving language datasets are infeasible for evaluating visual explanations of perceived risks: (i) a localized object is neither a risk nor an important object from the ego-perspective, but simply the object that the caption is referring to; and (ii) labels of these datasets are insufficient to address visual explanations in driving scenes. In~\cite{deruyttere2019talk2car}, captions are generated from a still image without considering temporal relationships against ego-vehicle. Also, per-pixel saliency in~\cite{kim2019grounding} is not easily mapped to important objects as driver gaze is often unrelated to the driving. In contrast, DRAMA accommodates ego-driver’s behavioral response to external events and explores video- and object-level semantic understanding obtained from video observations. 

Our annotation schema of question and answering is similar to the concept of VQA~\cite{antol2015vqa,krishna2017visual}. However, DRAMA specifically addresses visual explanations about driving risks associated with important objects, which cannot be discovered using existing dataset: (i) DRAMA facilitates in-depth analysis of driving scenes with annotations that contain domain-specific vocabularies like lane change, right lane, cut-in, traffic congestion, parked vehicle, etc.; and (ii) DRAMA is designed to show why the scenario is risky and what interactions are made as a free-form language description.\\
\section{DRAMA Dataset}
\begin{figure*}[t]
    \centering
    \includegraphics[width=0.85\textwidth]{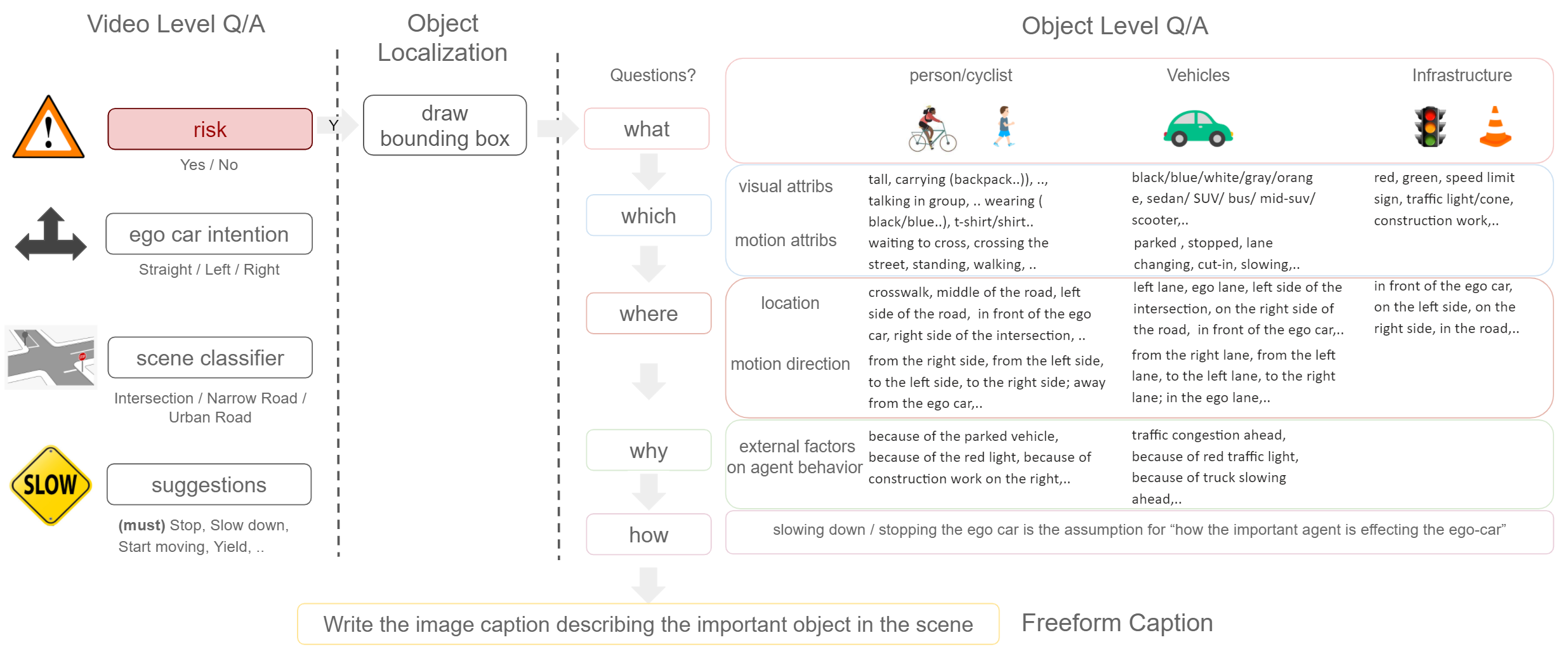}
    \caption{Annotation Schema and example labels for the DRAMA Dataset
    }
    \label{fig:drama_labels}
    \vspace{-0.3cm}
\end{figure*}

DRAMA is a dataset that provides visual reasoning of driving risks associated with important objects and that can be used to evaluate a range of visual reasoning capabilities from driving scenarios. We achieve this goal by generating video- and object-level questions and answers based on elementary operations (what, which, where, why, how) that represent object types, visual and motion attributes, object locations, or relationships of objects against the ego-vehicle. Figure~\ref{fig:drama_labels} provides an overview of the main components of DRAMA, which we considered for generating the annotations. 
In the following, we introduce our DRAMA dataset and describe its unique characteristics in detail. We first outline the data collection procedure including sensor configurations, raw data details, annotation schema, quality, and privacy in Section~\ref{sec:data_creation}. Next, we provide a statistical analysis of the dataset in Section~\ref{sec:data_statistics} and highlight the uniqueness by comparing with publicly available datasets 
in Section~\ref{sec:data_analysis}. Example scenarios of the DRAMA dataset are shown in Figure~\ref{fig:drama_data}.

\subsection{Dataset Creation}
\label{sec:data_creation}
We recorded 91 hours of videos from urban roads in Tokyo, Japan using SEKONIX SF332X-10X video camera (30HZ frame rate, $1928 \times 1280$ resolution and $60^\circ$ H-FOV) and GoPRO Hero 7 camera (60HZ frame rate, $2704 \times 1520$ resolution and $	118.2^\circ$ H-FOV). The videos are synchronized with the Controller Area Network (CAN) signals and Inertial Measurement Unit (IMU) information. We filtered out these videos based on the ego-driver's behavioral response to external situations or events, which activate braking of the vehicle. As a result, 17,785 interactive driving scenarios are created where each scenario is 2 seconds in duration. The annotators are requested to watch each scenario as observations and instructed to generate labels as described in the following paragraph.

The annotations of DRAMA are designed for targeting identification of important objects and generation of associated attributes and captions based on direct / indirect influences on the ego-car's motion. To obtain data in this form, we developed an annotation schema as follows:\vspace{0.4em}\\
\noindent
\textbf{Video-level Q/A}\;\; Questions are formatted to address risk and scene attributes as well as ego-behavioral functions such as ego-intention and behavior suggestions. Figure~\ref{fig:drama_labels} (left) illustrates these attributes in DRAMA. Some questions ask whether risks are perceived or what the driver's potential action is in order to ease the driving risk. The valid answers are closed-form such as a boolean format (\textit{i.e.}, yes or no) or single choice from multiple options (\textit{e.g.}, stop, slow down, start moving, ...).\vspace{0.4em}\\
\noindent
\textbf{Object-level Q/A}\;\; If the scenario is determined risky, the annotators proceed to a next step, object-level Q/A. It includes elementary operations (what, which, where, why, how) that can structure a question allowing a free-form or open-ended response. In addition to single choice answers, more complex reasoning is allowed to represent the high-level semantic understanding from observations. Figure~\ref{fig:drama_labels} (right) shows example answers that correspond to our elementary operations of visual reasoning such as querying object types (what), visual and motion attributes (which), location and motion direction (where), reasoning about risks or description of interactions (why), and effect of relationships (how).\\
For more details on privacy and quality consistency annotation procedure, please check supplementary material.
\noindent
\begin{figure*}[!h]
    \begin{subfigure}[b]{0.48\textwidth}
         \centering
         \includegraphics[width=\textwidth]{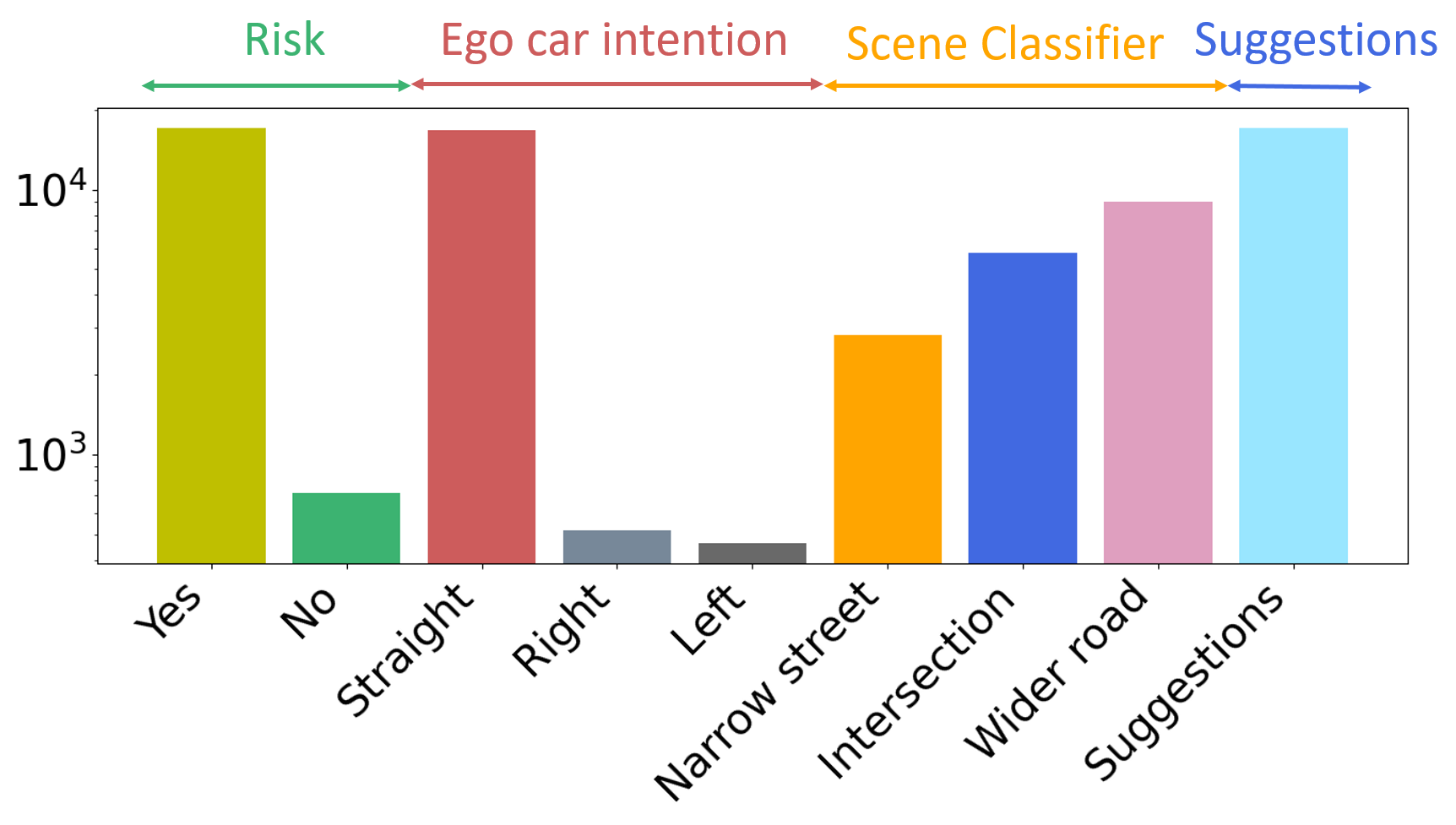}
         \caption{Distribution of video-level Q/A labels}
         \label{fig:scene_distribution}
     \end{subfigure}
     \begin{subfigure}[b]{0.46\textwidth}
         \centering
         \includegraphics[width=\textwidth]{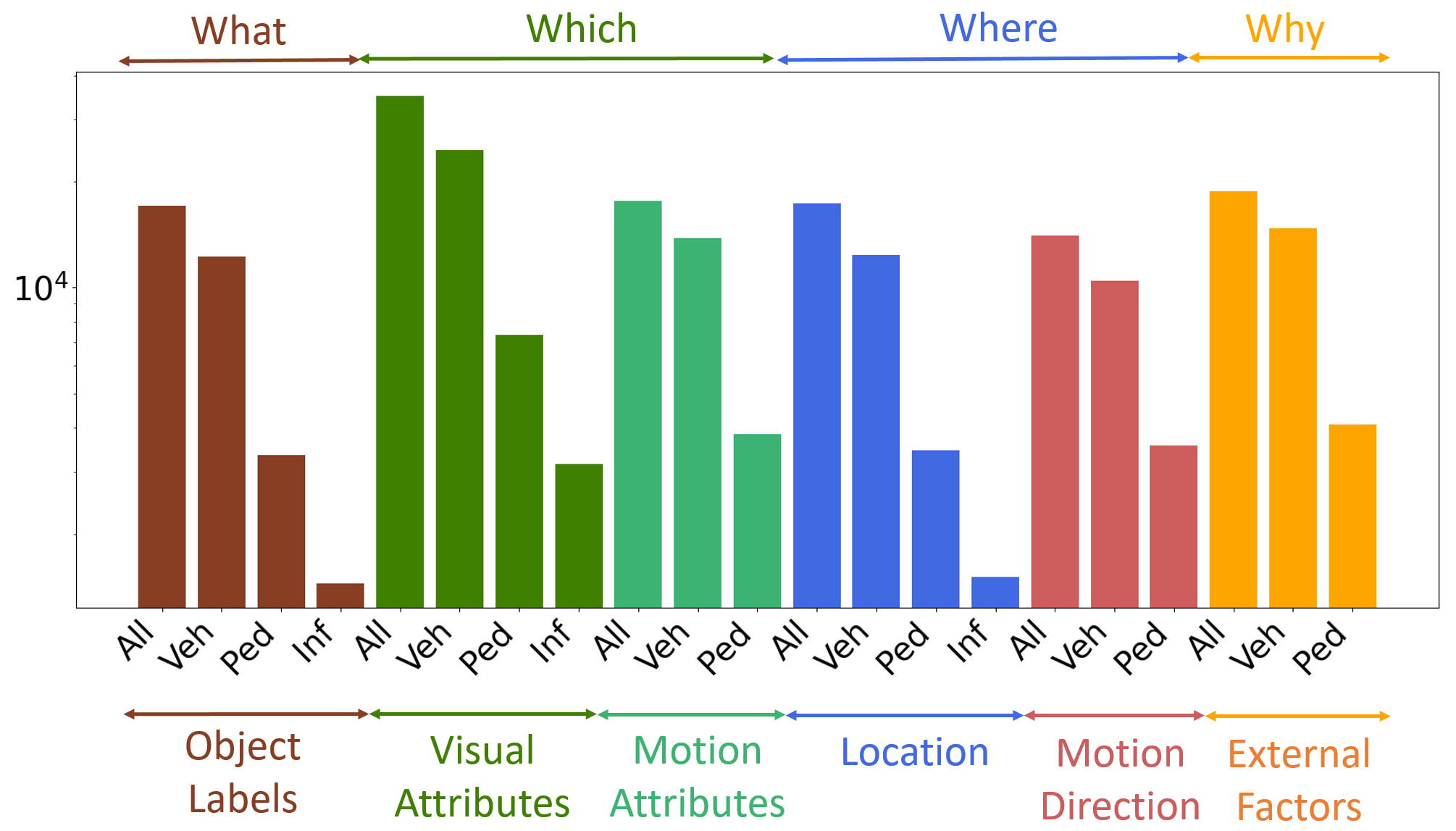}
         \caption{Distribution of object-level Q/A attributes}
         \label{fig:agent_level_distribution}
     \end{subfigure}
     \caption{Distribution of VQA attributes}
     \label{fig:attribute_distribution}
    \vspace{-0.4cm}
\end{figure*}
\subsection{Dataset Statistics}
\label{sec:data_statistics}
Figure~\ref{fig:scene_distribution} shows the distribution of labels obtained from video-level Q/A. Given the filtering of raw videos based on the activation of vehicle braking using CAN, majority of scenarios (17,066, 95.95\%) are labeled risky with the perception of important objects. 
The reasoning about risks was made during interactions of the important object when the ego-vehicle goes straight (94.5\%), turns left (2.6\%), or turns right (2.9\%). Each scenario was further categorized into three groups based on interactive environments: wider roads (51.5\%), intersections (32.6\%), and narrow streets (15.9\%). 

\begin{figure*}[!h]
\centering
\begin{subfigure}{0.46\textwidth}
  \centering
  \includegraphics[width=\linewidth]{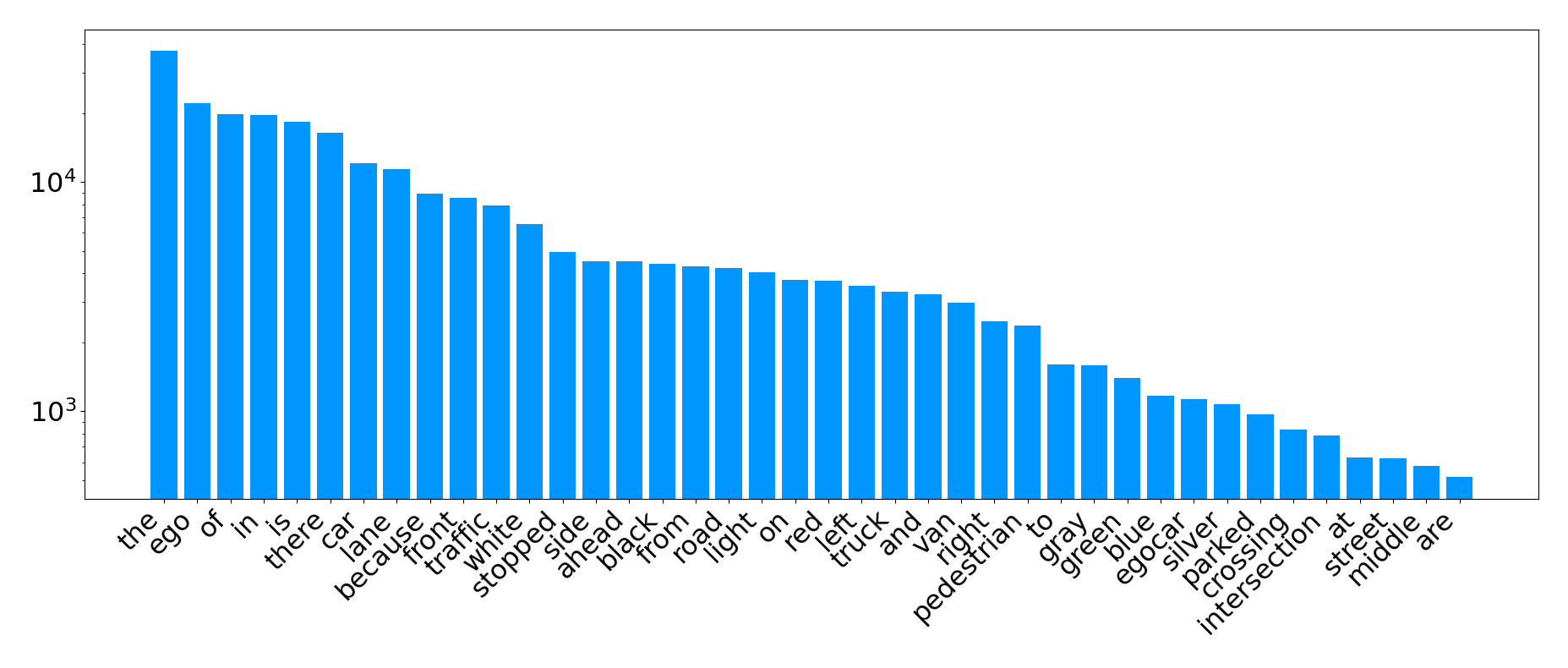}
  \caption{40 common words across all labels}
  \label{fig:common_words_dist}
\end{subfigure}
\begin{subfigure}{0.53\textwidth}
  \centering
  \includegraphics[width=\linewidth]{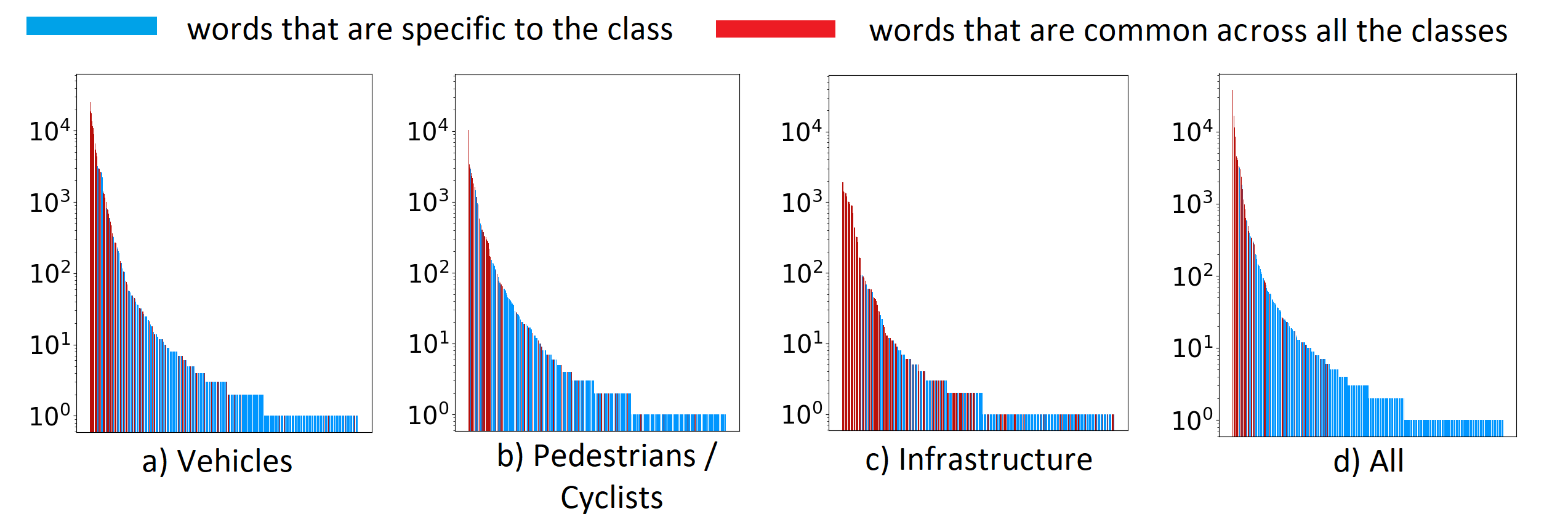}
  \caption{Word distribution per class}
  \label{fig:drama_dist_common_words}
\end{subfigure}
\caption{Statistics of words on the DRAMA dataset}
\vspace{-0.7cm}
\label{fig:words_dist}
\end{figure*}
Figure~\ref{fig:agent_level_distribution} presents the distribution of object-level Q/A attributes. The object that is recognized important was a vehicle (71.9\%), pedestrian or cyclist (19.6\%), and infrastructure (8.5\%). The visual attribute of these important objects can be generally answered with open-ended responses such as a red sedan vehicle, a tall person wearing blue shirt, red traffic light, etc., which results in 35,038 visual attributes from 17,066 objects. Similarly, motion attributes, location, motion directions, and reasoning about agent behaviors also have free-form values. 
In particular, the descriptions of reasoning include 992 unique words with total occurrences of 306,708 times. The number of unique words for vehicle, pedestrian or cyclist, and infrastructure is 533, 608 and 268, respectively. Also, 112 unique words are commonly shown across all object classes, which consist of colors (white, black, ...), locations (front, right, left), articles (the, an), prepositions (from, in, of, at), auxiliary verbs (is, are), adverbs (ahead, ...), nouns (street, road, car, ego, ...), conjunctions (because, and ), and actions (stopped, parked). The distribution of the words is shown in Figure~\ref{fig:words_dist}. More details are provided in the supplementary.

\subsection{Dataset Analysis}
\label{sec:data_analysis}
\begin{figure*}[t]
    \centering
    \includegraphics[width=0.95\textwidth]{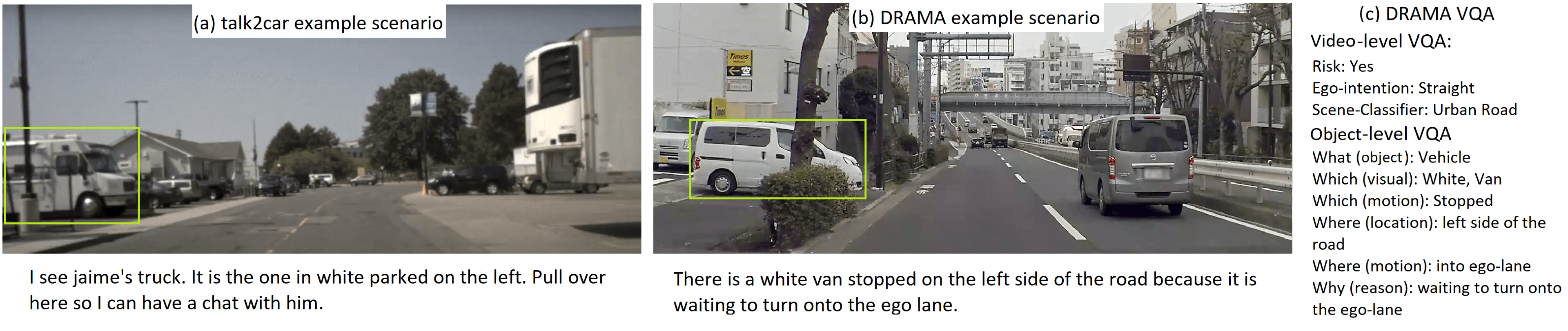}
    \caption{Comparison of an example scenario between (a) Talk2Car~\cite{deruyttere2019talk2car} and (b) our DRAMA dataset. Both scenarios annotate a white vehicle on the left as an object label (green bounding box) and provide an associated caption. (c) shows a snippet of video- and object-level attributes of DRAMA, corresponding to the scenario of (b).}
    \label{fig:comp_example}
\end{figure*}

\begin{table*}[t]
    \centering
    \resizebox{0.95\textwidth}{!}
    { 
    \begin{tabular}{l||c|c|c|c|c|c|c|c|c}
    
     & \multicolumn{4}{c|}{Risk Localization only} & \multicolumn{2}{c|}{Localization and Captioning}& \multicolumn{3}{c}{Captioning only} \\\hline
         
     Dataset & HDD~\cite{ramanishka2018toward} & EF~\cite{zeng2017agent} & SA~\cite{chan2016anticipating} & DoTA~\cite{yao2020and} & DRAMA (ours) & T2C~\cite{deruyttere2019talk2car} & HAD~\cite{kim2019grounding} & BDD-X~\cite{kim2018textual} &BDD-OIA~\cite{xu2020explainable} \\
         \hline\hline
         
    Risk localization &\cmark &\cmark & \cmark & \cmark & \cmark& no  &  no & no & no  \\
    
    Risk captions &no & no & no & no & \cmark& no   &\cmark & \cmark & \cmark\\     
         
    $\#$ of scenarios & 137 & 3,000 & 1,733 & 4677 & \textbf{17,785} & 9,217 &5,744 &6,984&11,303 \\
    
    $\#$ of captions & - &- & - & - & \underline{17,066}& 11,959 & 22,366& 26,534 & \textbf{27,265}\\

    Avg caption length& - &- & - & - & \textbf{17.97} & 11.01 & 11.05 & 8.90\textsuperscript{1} & 6.81\textsuperscript{1}\\

    VQA  & no & no & no & no & \cmark& no  & no& no & no \\

    $\#$ of questions &- & - & - & - & \textbf{77,639}& -  &-&- &-\\
    
    $\#$ of answers &- &- & - & - & \textbf{102,830}& -  & -&- &-\\
    
    Environment attribute & \cmark &no & no & no & \cmark & no  & no & no& no\\

    Reason / Why attribute & \cmark &no & no & no & \cmark & no & no & \cmark&  \cmark\\

    Scene-level risk attribute & \cmark & \cmark & \cmark & \cmark & \cmark& no  & no & no & no\\
    
    Freeform captions &no & no & no & no & \cmark& no   &\cmark & \cmark & no\\
    \hline
        
         \hline
    \end{tabular}
    }
    \caption{Comparison of DRAMA with public benchmark driving datasets.}
    \label{tab:comaparion_w_others}
\end{table*}
Table~\ref{tab:comaparion_w_others} shows the comparison of the DRAMA dataset to existing benchmarks. HDD~\cite{ramanishka2017top}, Epic Fail (EF)~\cite{zeng2017agent}, Street Accident (SA)~\cite{chan2016anticipating}, and DoTA~\cite{yao2020and} are object localization datasets that benchmark risk localization or anomaly detection. HAD~\cite{kim2019grounding},  BDD-X~\cite{kim2018textual}, BDD-OIA~\cite{xu2020explainable} are captioning datasets in driving that address advises and reasons for ego-vehicle actions in natural language descriptions.
Although these datasets are similar to DRAMA to some extent, we jointly address risk localization with visual reasoning of driving risks with a free-form language description.
Talk2Car (T2C)~\cite{deruyttere2019talk2car} is the image dataset that provides a caption associated with a bounding box. However, their object labels are not related to driving risks, and more importantly the caption is generated in a way that requests for an action to a car. In contrast, DRAMA provides important object labels with captions describing their risk from the ego-car perspective, while considering spatio-temporal relationships from videos. We directly compare our example scenario with that of \cite{deruyttere2019talk2car} in Figure~\ref{fig:comp_example}. Our caption labels are 1.4x more captions with 1.6x longer in length with respect to T2C. Unlike these benchmark datasets, DRAMA also delivers the functionality of visual question and answering to evaluate a range of visual reasoning capabilities from driving scenarios.

\section{Methodology}
\vspace{-0.2cm}

We introduce a model to address risk localization and its reasoning, which consist of an encoder and decoder as in Figure~\ref{fig:block_diagram1}. The details of model architecture are shown in the supplementary material.
\begin{figure*}[t]
\centering
\includegraphics[width=0.85\linewidth]{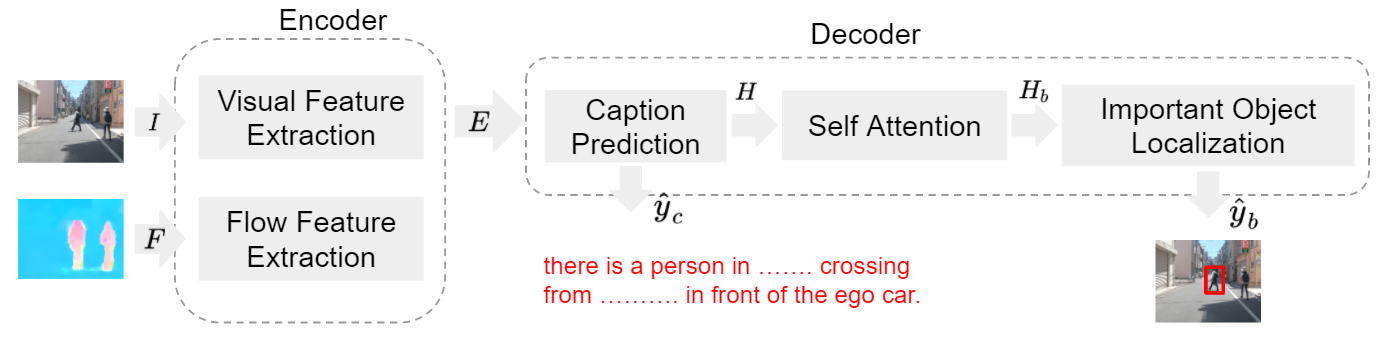}
\caption{DRAMA Architecture: the encoder and decoder modules}
\label{fig:block_diagram1}
\end{figure*}

\textbf{Encoder}\;\; Given an image $I_t$ and an optical flow image $F_t$ at time step $t$, the encoder extracts visual and optical flow features. The optical flow image $F_t$ is generated by Flownet 2.0~\cite{ilg2017flownet} with two image frames $I_t$ and $I_{t-1}$ to capture the motion of objects and scene. For notational brevity, we drop $t$ in the following sections. 
To encode visual feature and optical flow feature, we use a ResNet-101 model~\cite{he2016deep} pre-trained on Imagenet~\cite{krizhevsky2012imagenet} as a backbone network. Both features are extracted from two layers prior to the original ResNet-101 structure, and we add the adaptive average pooling layer to get a fixed embedding size of 14x14x2048 as outputs. We further concatenate these two features to generate the output feature $E$.

\textbf{Decoder}\;\; The decoder is trained to find a bounding box ($\hat{y}_{b}$) as an important object and to generate a corresponding caption description ($\hat{y}_{c}$). In our proposed model, we first predict a caption $\hat{y}_{c}$ using an LSTM-based recurrent model. The hidden states of LSTM, $H = [h_1, \dots, h_M]$, are used as the embedding for the self-attention~\cite{Vaswani2017selfattention}. The resulting new feature $H_{b} = [h_{b1}, \dots h_{bM}]$ is obtained by $h_{bi} = self\;attention (H W_i^Q, H W_i^K, H W_i^V)$, where $W^Q$, $W^K$ and $W^V$ are parameter matrices for queries, keys and values, respectively. Lastly, we find a bounding box $\hat{y}_{b}$ using an important object localization model. In the experiments, the caption prediction model follows the decoder architecture of SAT~\cite{xu2015show}, which is a LSTM model with the soft-attention mechanism. Also, we used a basic multi-layer perceptron (MLP) for the important object localization model.

\textbf{Loss Function}\;\; The loss function for joint risk localization and reasoning is defined as follows:
\begin{align}
    L_b &= \sum_{j=1}^{N} \norm{\hat{y}_{b}^{(j)} - y_{b}^{(j)}}_2; \label{eq:loss_cap_b} \\
    L_c &= \sum_{j=1}^{N} \left[ \sum^{M}_{k=1} ce \left(\hat{y}_{ck}^{(j)}, \; y_{ck}^{(j)} \right) + \lambda \sum_{i=1}^{P} \left( 1 - \sum_{k=1}^{M} \alpha_{ki}^{(j)} \right)^2 \right], \label{eq:loss_cap_c}
\end{align}
where $^{(j)}$ denotes the $j$-th data and the right term of $L_{c}$ is a doubly stochastic attention regularizer adapted from SAT~\cite{xu2015show}. $\alpha_{ki}$ denotes an element in the attention map in the caption prediction, $M$ and $P$ are the caption length and the number of pixels in the attention map, respectively, and $\lambda$ is a hyperparameter. Then, we use a multi-task loss~\cite{Roberto2018multitask} for combining $L_b$ and $L_c$, which results in
\begin{equation}
    L_{total} = \frac{1}{\sigma^2_1} L_{b}+\frac{1}{\sigma^2_2} L_{c}+ \log(\sigma_1 \sigma_2),
\label{eq:loss_tot}
\end{equation}
where $\sigma_1$ and $\sigma_2$ denote the relative weights of the losses $L_b$ and $L_{c}$, respectively. The multi-task loss function considers the homoscedastic uncertainty of each task by training the weights $\sigma_1$ and $\sigma_2$.
\vspace{-0.2cm}
\section{Experiments}
\subsection{Model Comparison}
\blfootnote{\textsuperscript{1}reported average caption length is without ego action for fair comparison with our dataset, otherwise it is 13.92 for BDD-X and 8.36 for BDD-OIA.} We evaluate the performance of our model on the new DRAMA dataset, by comparing with various baseline models. For the fair comparison, all baselines use the same backbone networks; pre-trained ResNet-101~\cite{he2016deep} for the encoder, the SAT~\cite{xu2015show} structure for caption prediction (CP), and MLP for important object localization (IOL). Since our model is conditioned on captioning for localization, we named LCP (localization with captioning prior). Our baseline models are described as follows:

\vspace{0.2cm}
\textbf{ResNet-101}~\cite{he2016deep} \;\; This is a strong baseline model that identifies an important object only, which is trained with Equation~\eqref{eq:loss_cap_b}. As mentioned earlier, our structure for the localization task is built from the original ResNet model by replacing its last two layers by adaptive average pooling and MLPs as the decoder. Therefore, this model provides an insight on the efficacy of an additional optical flow input as well as our total loss function in Equation~\eqref{eq:loss_tot}.

\vspace{0.2cm}
\textbf{SAT}~\cite{xu2015show} \;\; This baseline model only generates a caption description. The model takes both an image and optical flow as input and have the loss function of Equation~\eqref{eq:loss_cap_c}. We compare this baseline with LCP to demonstrate our captioning capability under joint training of risk localization and reasoning.

\vspace{0.2cm}
\textbf{Independent captioning and localization (ICL)} \;\; The ICL model simultaneously finds the important object $\hat{y}_b$ and caption description $\hat{y}_c$ independently from one another. This model is trained using the dual task loss function in Equation~\eqref{eq:loss_tot}, same as our LCP model.

\vspace{0.2cm}
\textbf{Captioning with localization prior (CLP)} \;\; In opposition to LCP, this baseline model has the decoder architecture that generates captioning using a localization prior. We add a region of interest (ROI) pooling to improve the model's captioning capability. The ROI pooling produces the important object feature $E_a$ from the encoder output $E$. Then, we concatenate $E_a$ with the output of IOL, and the concatenated features are used as the input of the caption generation. This model is similar to~\cite{johnson2016densecap} with a small variation of skip connection of the encoder output $E$, to get global information. This model is also trained using the dual task loss function in Equation~\eqref{eq:loss_tot}.

We evaluate the effectiveness of optical flow information by creating additional baselines using LCP, ICL and CLP models without optical flow stream. Furthermore, we generate additional LCP without the multi-task loss function, which validates our effort of balancing the multi-task training.

\subsection{Experimental Setting and Metrics}
\label{sec:experimental_setting}
We split the DRAMA dataset into 70\% train, 15\% validation and 15\% test. The model parameters are chosen on the validation set at training time, and we report the performance of this model tested on the test set. The captioning performance is evaluated using the standard metrics such as BLEU-1 (B1)~\cite{papineni2002bleu}, BLEU-4 (B4)~\cite{papineni2002bleu}, METEOR (M)~\cite{banerjee2005meteor}, ROGUE (R)~\cite{lin2004rouge}, CIDER (C)~\cite{vedantam2015cider}, and SPICE (S)~\cite{anderson2016spice}. We use Mean-IOU (Intersection Over Union) and accuracy for IOU$>$0.5 for the evaluation of important agent localization. IOU is a standard metric in object detection, dividing the area of overlap by the area of union between the predicted bounding box and the ground-truth bounding box. We provide other experimental settings in the supplementary.


\begin{figure*}[t]
    \centering
    \includegraphics[width=0.8\textwidth]{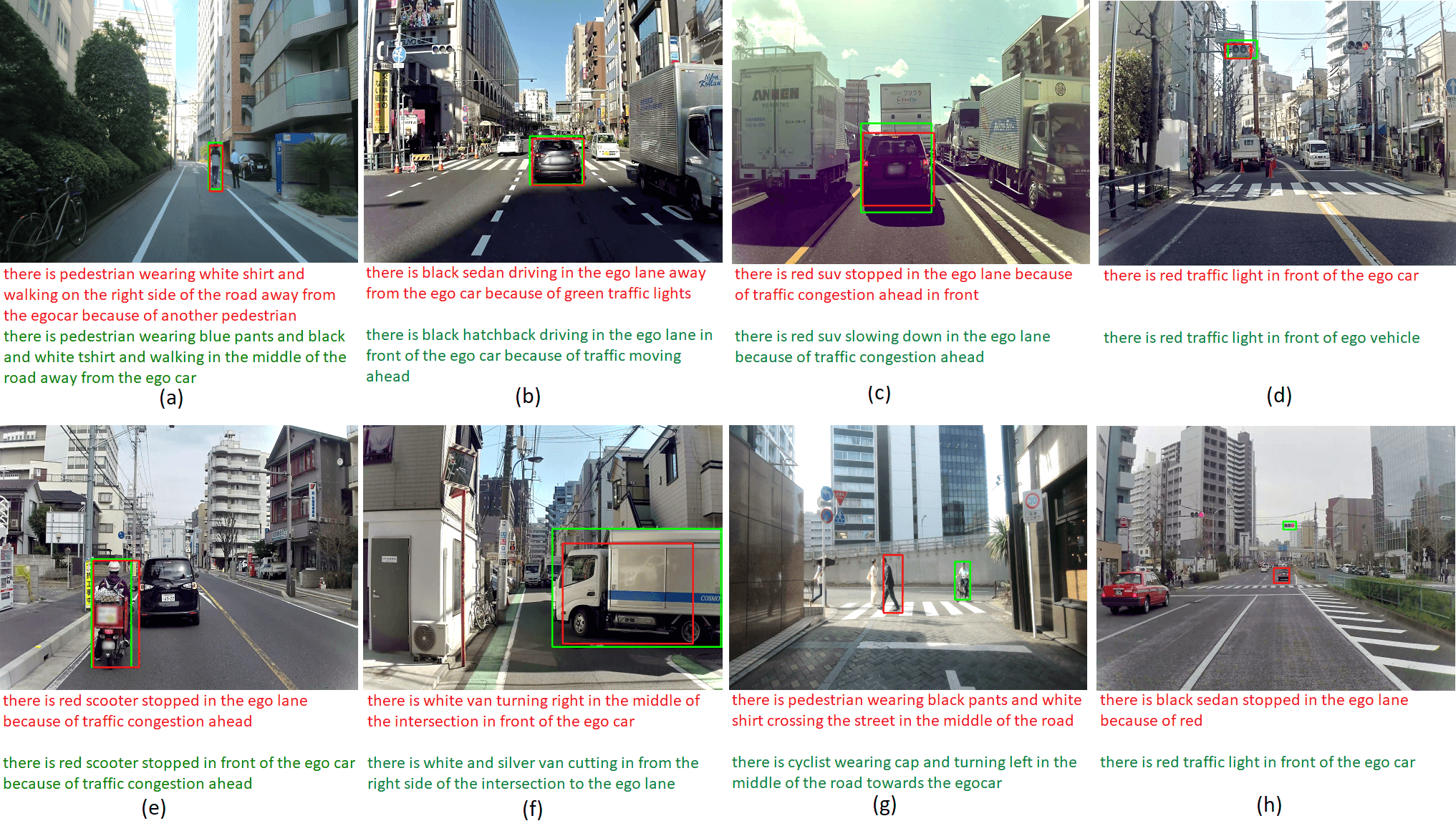}
    \caption{Qualitative results of the DRAMA model: our prediction in red and ground truth in green. More detailed success and failure cases are discussed in the supplementary material.}
    \label{fig:qualitative}
\end{figure*}

\subsection{Results}
\label{sec:results}
\begin{table*}[!h]
\centering
\resizebox{0.7\textwidth}{!}
{ 
\begin{tabular}{ l||c|c|c|c|c|c|c|c }
\hline
& \multicolumn{6}{c|}{Captioning} & \multicolumn{2}{c}{Important Object Localization} \\
\hline
Method & B1 $\uparrow$ & B4 $\uparrow$ & M$\uparrow$ & R $\uparrow$ & C $\uparrow$ & S $\uparrow$ & Mean-IOU $\uparrow$ & Acc (IOU$>$0.5) $\uparrow$ \\
\hline\hline
ResNet-101~\cite{he2016deep} & - & - & - & - & - & - & 0.600 & 0.670 \\
SAT~\cite{xu2015show} & 0.740 & 0.531 & 0.386 & 0.694 & 3.583 & 0.544 & - & - \\
\hline
ICL without OF & 0.731 & 0.518 & 0.377 & 0.679 & 3.468 & 0.522 & 0.553 & 0.617 \\
ICL & 0.739 & 0.539 & 0.385 & 0.693 & 3.678 & 0.534 & 0.533 & 0.593 \\
CLP without OF ~\cite{johnson2016densecap}& 0.744 & 0.544 & 0.386 & 0.700 & 3.582 & 0.524 & 0.518 & 0.587 \\
CLP & \textbf{0.744} & 0.546 & 0.388 & 0.701 & 3.670 & 0.540 & 0.520 & 0.581 \\
\hline
LCP without OF & 0.721 & 0.520 & 0.379 & 0.688 & 3.501 & 0.536 & 0.597 & 0.667 \\
LCP without DTL & 0.740 & 0.543 & 0.389 & \textbf{0.703} & 3.668 & 0.548 & 0.569 & 0.640 \\
LCP (\textbf{Ours}) & 0.739 & \textbf{0.547} & \textbf{0.391} & 0.700 & \textbf{3.724} & \textbf{0.560} & \textbf{0.614} & \textbf{0.684} \\
\hline
\end{tabular}
}
 \caption{Quantitative evaluation comparing our model LCP (localization with captioning prior) with ICL (independent captioning and localization) and CLP (captioning with localization prior). OF and DTL denotes optical flow and dual task loss function, respectively.}
\vspace{-0.5cm}
 \label{tab:results_table}
\end{table*}
\begin{table}[!h]
\centering
\resizebox{0.48\textwidth}{!}
{ 
\begin{tabular}{ l||c|c|c|c}
\hline
& \multicolumn{2}{c|}{ C $\uparrow$} &\multicolumn{2}{c}{ Mean-IOU $\uparrow$} \\
\hline
Object & SAT & LCP (ours) & ResNet-101 & LCP (ours) \\ 
\hline\hline
Vehicle & 3.588 & \textbf{3.717} & 0.714 & \textbf{0.737} \\
Pedestrian & 2.100 & \textbf{2.360} & 0.266 & \textbf{0.286} \\
Cyclist & 1.630 & \textbf{1.912} & \textbf{0.276} & 0.271 \\
Infrastructure & 6.216&\textbf{6.480} & 0.091 & \textbf{0.109} \\
\hline
\end{tabular}
}
\caption{LCP is compared with SAT~\cite{xu2015show} and ResNet-101~\cite{he2016deep} with respect to object types.}
 \vspace{-0.8cm}
 \label{tab:results_table2}
\end{table}
\textbf{Quantitative comparison} \;\; Table~\ref{tab:results_table} shows the quantitative results of our model and baseline models on the DRAMA dataset. First, we can see that adding the optical flow information is a positive effect in captioning across all the decoder baselines including LCP, ICL and CLP. The reason is the words related to motion such as moving, stopped, etc. can be captured by the optical flow information, but not by a single image. Since the video dataset can only use optical flow images, our DRAMA dataset has more benefits than the image dataset such as the Talk2Car~\cite{deruyttere2019talk2car} dataset. Next, our LCP model performs better than ICL and CLP baselines in important object identification. It makes sense that LCP performs well in the localization due to an additional self-attention module. Note that the captioning performance of LCP is higher or similar to that of CLP and better than those of ICL and SAT. It implies that better localization derived by self-attention influenced to finding a better multi-task weighting in our loss function. Although the ROI pooling in the CLP model improves the captioning comparing to ICL and SAT, it seems to negatively affect to risk localization. Additionally, it shows dual task loss helps to improve the overall performance of LCP (LCP vs LCP without DTL).

We compare our model to the single task baselines, SAT and ResNet-101, in detail with respect to the type of objects. We use the CIDER and Mean-IOU metrics for the comparison. Table~\ref{tab:results_table2} shows that our model performs better than SAT for all types of objects in the captioning. Also, important object identification of our method is better than that of the ResNet-101 model for most of object types. It demonstrates that both tasks are highly correlated and can benefit being tackled together.

\textbf{Qualitative comparison} \;\; Figure~\ref{fig:qualitative} shows the ground truth and our prediction in the cases where the important objects are pedestrians (a,g), vehicles (b,c,e,f) and infrastructure (d,h). Figure~\ref{fig:qualitative}a and \ref{fig:qualitative}b show the model learns a good reasoning such as ``because of another pedestrian'' and ``because of green traffic light'' even though the ground truth does not have. In Figure~\ref{fig:qualitative}c and \ref{fig:qualitative}e, the reasoning ``because of traffic congestion ahead'' is correctly predicted by our model. In Figure~\ref{fig:qualitative}f, the prediction correctly describes the object in a different way than the ground truth, \textit{e.g.}, ``turning right in the middle of the intersection'' vs. ``cutting in from the right side of the intersection.'' Figure~\ref{fig:qualitative}g and \ref{fig:qualitative}h show our model fails to find the ground-truth important object and caption description. Although our model identifies the object closer to the ego-vehicle rather than the object heading towards the ego-vehicle in Figure~\ref{fig:qualitative}g, it is also considered as a driving risk observed in the scene. We believe such an issue can be addressed by generating multi-modal outputs, which is our future research direction. In Figure~\ref{fig:qualitative}h, the red traffic light is closer and more important than the vehicle, but our model identifies the vehicle mainly because the model is incapable of understanding the physical distance between objects. Extension to incorporate such geometric information is our next plan. 

\section{Conclusion}
We introduced a new research direction of joint risk localization in driving scenes and its explanations as a natural language description. To address this problem, we collected a novel driving dataset, DRAMA, which is annotated with video- and object level questions on important objects as well as an interpretation of the scene with respect to the interaction observed from the ego-driver's perspective. As a result, DRAMA facilitates the evaluation of a range of visual explanation capabilities from driving scenarios. We designed various models for the purpose of benchmarking jointly supervised frameworks that localize a risk and its linguistic explanations as a natural language description. Extensive ablative study on different architectures demonstrated that important object localization using captioning as a prior showed the improved performance against other baseline models.

{\small
\bibliographystyle{ieee_fullname}
\bibliography{egbib}
}
\begin{figure*}[t]
\begin{center}
 \includegraphics[width=0.9\textwidth]{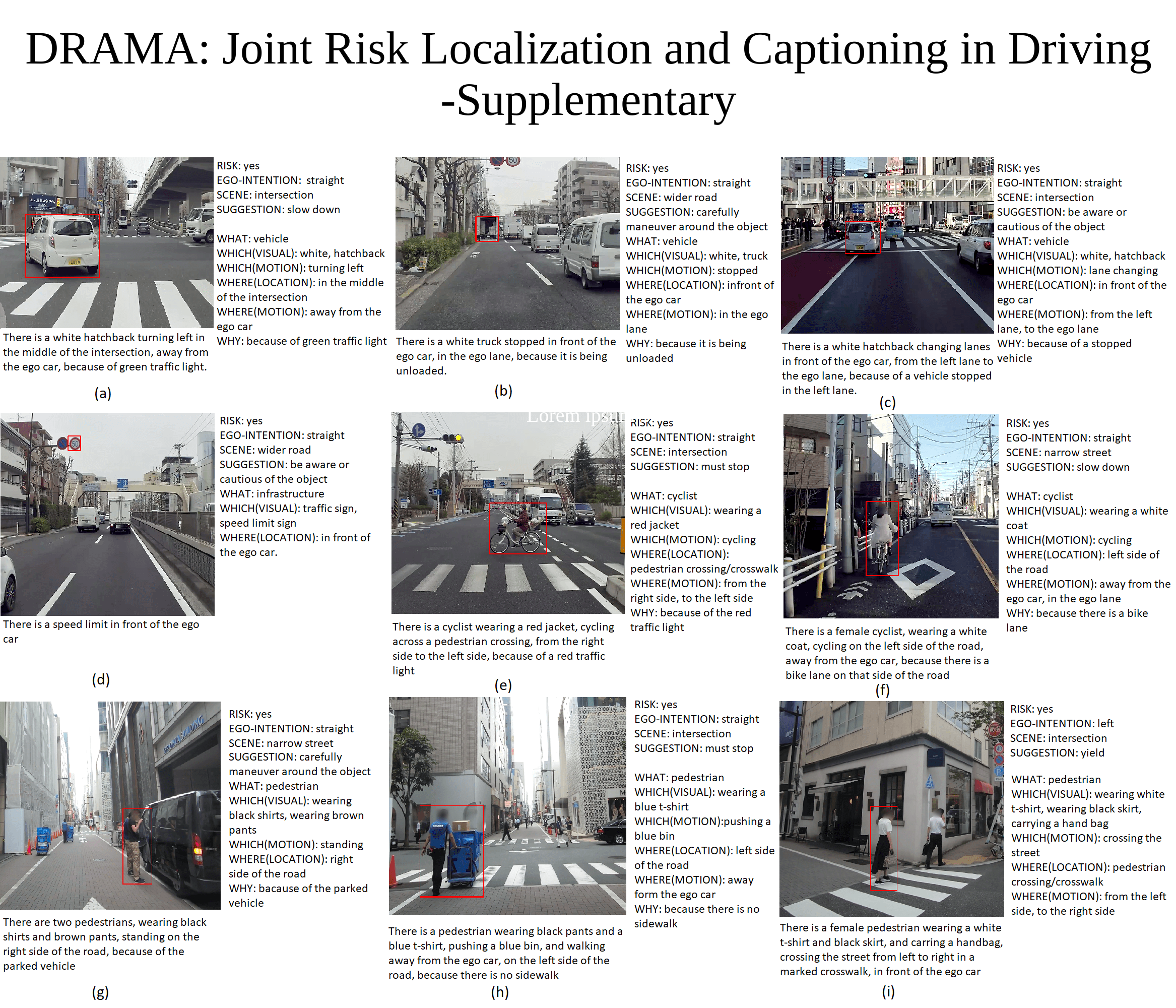}
\captionof{figure}{Example scenarios of the DRAMA dataset. Risks are perceived while interacting with different vehicles in (a-c), infrastructure in (d), cyclists (e,f), and pedestrians(g-i).}
\label{fig:data_scenarios}
\end{center}
\end{figure*}
\newpage
\phantom{a}


\begin{table*}[!t]
\centering
\resizebox{0.99\textwidth}{!}
{\begin{tabular}{l||c|c|c|c|c|c|c|c|c}
&\multicolumn{5}{c|}{Object-level Q/A Attributes}
& \multicolumn{2}{c|}{Bounding Box}
& \multicolumn{2}{c}{Captioning} \\ \hline
Class& visual attr & motion attr & location & motion direc & ext fact& no of & normalized & unique & total \\
(what) & (which) & (which) & (where) & (where) & (why) & boxes & mean box size & words & occurrences \\ \hline \hline
Veh & 24,546 & 13,829 & 12,350 & 10,462 & 14,712 & 12,273 & 0.0846 & 533 & 214,511  \\
Ped/Cyc & 7,338 & 3,829 & 3,454 & 3,571 & 4,080 & 3,344 & 0.0405 & 608 & 75,576  \\
Inf & 3,154 & - & 1,505 & - & - & 1,449 & 0.0102 & 268 & 16,616  \\ \hline
All & 35,038 & 17,658 & 17,309 & 14,033 & 18,792 & 17,066 & 0.0697 & 992 & 306,708 \\ \hline
\end{tabular}}
\vspace{0.25cm}
\caption{The statistics of obect-level Q/A attributes, bounding box, and captions.}
\vspace{-0.25cm}
\label{tab:data_stats_table}
\end{table*}
\section{Additional Analysis of DRAMA}
\subsection{Example Scenarios}
We visual various scenarios of our DRAMA dataset in Figure~\ref{fig:data_scenarios}. Each scenario includes video- and agent-level attributes annotated from sequence and motion observations. 
Some actions of vehicles such as turning left, stopped, lane changing are shown in Figure~\ref{fig:data_scenarios}a-\ref{fig:data_scenarios}c respectively in various interactive environments (intersection, wide and narrow road) where the reasoning of the important object is described by `because of traffic light' in \ref{fig:data_scenarios}a, `because it is being unloaded' in \ref{fig:data_scenarios}b, ` because of a stopped vehicle' in \ref{fig:data_scenarios}c, respectively. 
We show cyclists interacting with the ego car at `intersections or crosswalks' in \ref{fig:data_scenarios}e and `narrow roads or driving in ego lane' \ref{fig:data_scenarios}f. Pedestrians are also identified as important agents interacting with the ego-car while performing different actions such as `standing', `pushing',`crossing the street' as shown in Figure~\ref{fig:data_scenarios}g-\ref{fig:data_scenarios}i. Their importance is reasoned by behaviors like `because of parked vehicle' in \ref{fig:data_scenarios}g and `because there is no side walk' in \ref{fig:data_scenarios}h. In case of Figure~\ref{fig:data_scenarios}i, the risk is caused by internal stimuli rather than affected by external influences, so WHY is not annotated in this scenario. Similarly, the infrastructure only has visual and location questions in Figure~\ref{fig:data_scenarios}d.

\subsection{Statistics}
We show the data statistics of DRAMA in Table~\ref{tab:data_stats_table}. Note that video-level Q/A attributes are not included in the table as the annotated number is same as the number of scenarios (17,785). 

The visual attributes for vehicles are 2 per a bounding box on average (PBA) as they can be described by the color and type of the vehicle. Whereas, pedestrians/cyclists has 2.19 PBA, which indicates these agents have more descriptive representations. The visual attributes for the infrastructure usually include it's state and name, PBA is measured at 2.17. The motion attributes for vehicles is 1.12 PBA, as it describes the motion state like stopped, parked, etc. Similarly for pedestrians/cyclists, it is 1.14 PBA. The location attributes for vehicles, pedestrians/cyclists, and infrastructure are respectively 1.00, 1.03, 1.06 PBA. The motion direction for pedestrian/cyclists is  1.06 PBA, which is higher than that of vehicles (0.85 PBA). This is mainly because most of the influence happens when pedestrians/cyclists cross the road or cyclists move slowly away/towards the ego-car, which requires more descriptions. On the other hand, in case of vehicles, they are identified important when they slow down (or braking) in front of the ego-vehicle without describing the direction. The external factors or second level of reasoning (reasoning for the important agent's behavior) for vehicles is 1.20 PBA and pedestrians/cyclists 1.22 PBA. The infrastructure neither moves nor gets influenced by other external factors, so the last two columns in agent level VQA attributes in Table~\ref{tab:data_stats_table}. 

The bounding box proportions across different object categories are also shown in Table~\ref{tab:data_stats_table}. The vehicle bounding boxes are the most dominant ones 12,273 (71.91\%) of the all the boxes. The normalized mean bounding boxes size (NMBS) of the vehicle category is 8\% from the entire image as they are often very close to the ego-vehicle.  There are 2,909 pedestrians and 435 cyclists, which is 19.19\% of all the bounding boxes in the DRAMA dataset. The NMBS of this object category is 4\% of the image. We can infer that the ego-car slows or stops from a distant location when their sizes are small from the egocentric perspective. The bounding boxes of infrastructure are the least percentage 1,449 (8.49\%), with NMBS of 1.02\%. Most of infrastructure is construction cones, traffic lights, and traffic signs, which cover only a small portion of the image. Resulting in poorer performance for object localization for infrastructure compared to others in Table 3 can be partially described from their NMBS.

In total, the vocabulary used in the dataset consists of 992 unique words, with total occurrences of 306,708. The vocabulary size of vehicles is 533, with total occurrences of 214,511 (69.94\%). The vocabulary size of pedestrians and cyclists is 608 with total occurrences of 75,576 (24.64\%). The vocabulary size of infrastructure is 268 with total occurrences of 16,616 (5.41\%). The proportions of the words are proportional to number of those objects present in the dataset. The common unique words used in captioning across all classes are 112 with total occurrences of 265,922 (86.7\%). Additional information can be found in the main manuscript in Section 3.2 and Figure 4.

\begin{table*}[!t]
\centering\small
\begin{tabular}{ c|l|c|c}
 &Layer & Kernal shape & Output shape\\
\hline\hline
  \multicolumn{1}{c}{Flow/Visual Encoder}&\multicolumn{3}{c}{}\\\hline
0&resnet.Conv2d\_0  & [3, 64, 7, 7] & [1, 64, 370, 500]\\
1&resnet.BatchNorm2d\_1 & [64] & [1, 64, 370, 500]\\
2&resnet.ReLU\_2 & -  & [1, 64, 370, 500]\\
3&resnet.MaxPool2d\_3 & - &  [1, 64, 185, 250]\\\hline
..&\multicolumn{3}{c}{... (resnet layers)}\\\hline
306&resnet.7.2.Conv2d\_conv3 & [512, 2048, 1, 1]  & [1, 2048, 24, 32]\\
307&resnet.7.2.BatchNorm2d\_bn3 &[2048] & [1, 2048, 24, 32]\\
308&resnet.7.2.ReLU\_relu & - & [1, 2048, 24, 32]\\
309&AdaptiveAvgPool2d\_adaptive\_pool &- & [1, 2048, 14, 14]\\\hline
\multicolumn{1}{c}{Decoder}&\multicolumn{3}{c}{}\\\hline
310(a)&concat.visual\_flow &-& [1, 4096, 14, 14]\\
310(b)&module.Embedding\_embedding &[512, W]&[1, T, 512]\\
311(a)&Linear\_init\_h &  [4096, 512]&[1, 512]\\
311(b)&Linear\_init\_c &  [4096, 512]&[1, 512]\\
312&LSTM\_cell\_with\_attention&-&[1, 512]\\
313&word\_dropout\_0&-&[1, 512]\\
314&word\_fc&[512,W]&[1, W]\\
315&word\_softmax&-&[1, W]\\\hline
316&\multicolumn{3}{c}{rollout LSTM for M times, from step 312 to 315}\\\hline
317&\multicolumn{2}{c|}{Self-attention Step, Sec 4}&[1, 512]\\\hline
318&bbox\_predictor.Linear\_0&[512, 4]&[1, 4]\\
319&bbox\_predictor.Sigmoid\_1&-&[1,4]\\\hline
\hline
\end{tabular}
 \caption{DRAMA summary, for input image of size (3,740,1000), batch size as 1, vocabulary size (W) as 989 and max sentence length (M) as 50}
  \label{tbl:model_summary}
\end{table*}

\subsection{Privacy}
An open-source tool, Anonymizer~\cite{understand-ai} is used to anonymize faces and vehicle number plates as a coarse annotation step. The annotators find unblurred faces or license plates and manually blurred them. 

\subsection{Question Representation}
For a comprehensive annotation of DRAMA, we generate a set of questions that can yield various forms of answers from closed-ended such as boolean (yes or no) to open-ended such as `pedestrian wearing black pants and white shirt' or `orange traffic cones on the right'. We used elementary operations (\textit{i.e.}, what, which, where, why, and how) to reason about ego-driver’s behavioral response to perceived risk that activates braking of the vehicle. The questions for querying video- and object-level attributes are as follows.

Video-level questions:
\begin{itemize}
\item Is there any `\textit{risk}' in the scene?
\item What is the `\textit{intention}' of the ego-car?
\item What is this `\textit{scene}'?
\item What `\textit{suggestions}' do you give for the driver to avoid risk?
\end{itemize}

If risks are perceived in the video, questions regarding the important agent are:
\begin{itemize}

 \item `\textit{What}' agent is it?
 \item If Pedestrian/Cyclist, `\textit{Which}' visual attributes best describe the important agent?
 \item If Vehicles, `\textit{Which}' type of the vehicle is it? (visual attributes)
 \item If Vehicles, `\textit{Which}' color is the vehicle? (visual attributes)
 \item If Infrastructure, `\textit{Which}' name is the infrastructure called? (visual attributes)
 \item If Infrastructure, `\textit{Which}' state is it in? (visual attributes)
 \item  `\textit{Which}' motion attributes best describe the agent?
 \item  `\textit{Where}' is the location of the agent?
 \item  `\textit{Where}' is the agent moving (from or to)?
 \item  `\textit{Why}' is the object behaving the way it is?
\end{itemize}

\section{Implementation Details}
\subsection{Bounding Box Parameterization}
The bounding box is notated using \{cx,cy,l,w\}, where (cx,cy) is the center coordinate and (l,w) are the dimension of the bounding box. They are normalized using the size of the image (L,W). During inference, we use a sigmoid activation function to find the values in the range of [0,1] and project it back to the original pixel coordinates to evaluate the output using the metrics (\textit{i.e.}, Mean-IOU and Accuracy).

\begin{figure*}[!t]
    \begin{subfigure}[]{0.48\textwidth}
         \centering
         \includegraphics[width=\textwidth]{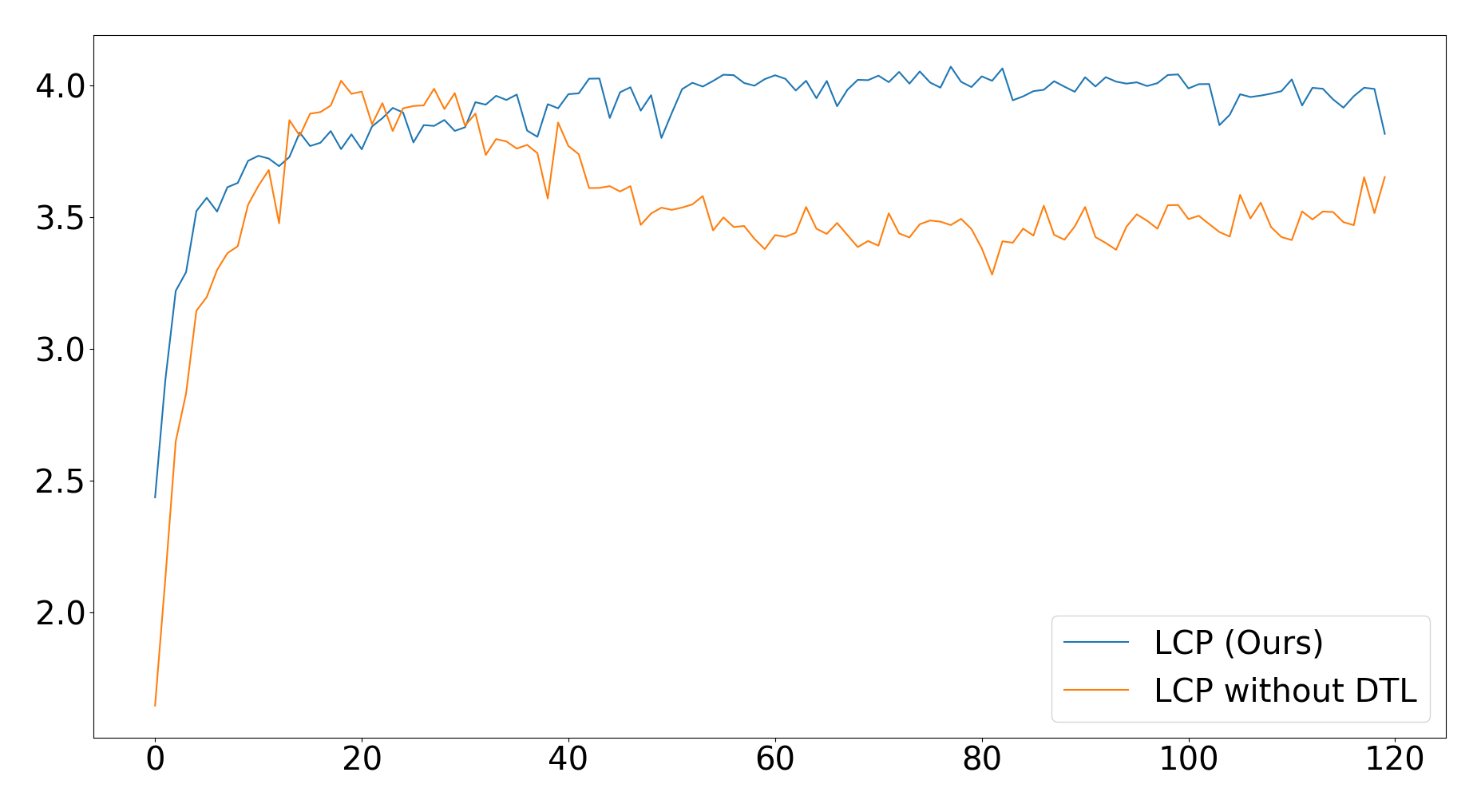}
         \caption{}
         \label{fig:val_cider}
     \end{subfigure}
     \hfill
     \begin{subfigure}[]{0.48\textwidth}
         \centering
         \includegraphics[width=\textwidth]{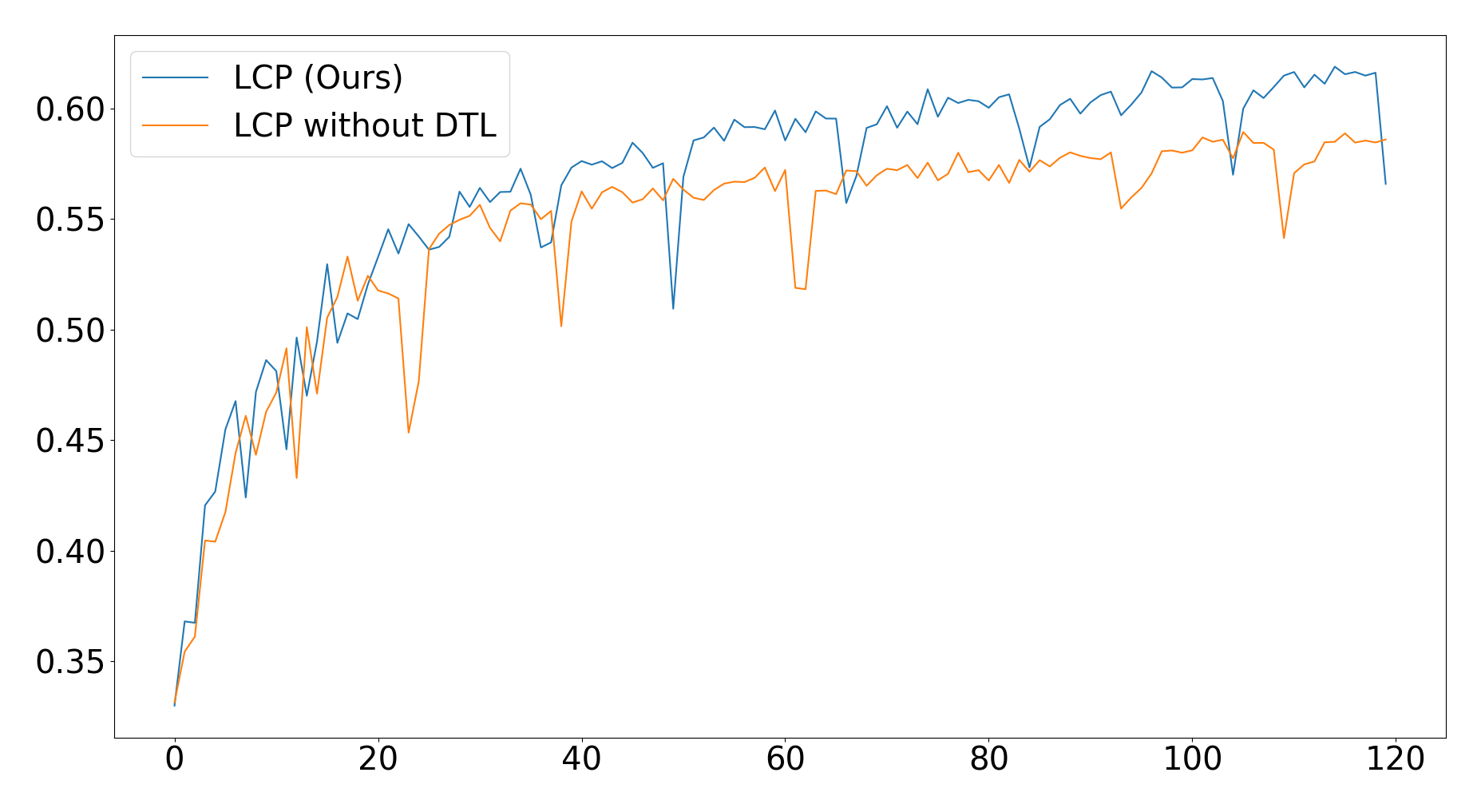}
        \caption{}
         \label{fig:val_iou}
     \end{subfigure}
     \caption{(a) CIDEr score on validation set during training and (b) Mean-IOU score on validation set during training.}
     \label{fig:val_curves}
     \vspace{-0.25cm}
\end{figure*}
\begin{table*}[t!]
    \centering
    \resizebox{0.98\textwidth}{!}
    { 
    \begin{tabular}{l||c|c|c|c|c|c|c|c}
    Metrics&B1&B4&M&R&C&S&Mean-IOU&Acc(IOU>0.5)\\\hline
    LCP (Ours) &0.739$\pm$0.0065&0.547$\pm$0.011&0.391$\pm$0.0045& 0.700$\pm$0.004&3.724$\pm$0.0765&0.560 $\pm$0.009&0.614$\pm$0.0075& 0.684$\pm$0.0115\\
    \end{tabular}
    }
    \vspace{0.25cm}
    \caption{Table showing standard deviations of our model for each metrics}
    \vspace{-0.25cm}
    \label{tab:std_deviation}
\end{table*}
\begin{figure}[h]
\centering
\includegraphics[width=\linewidth]{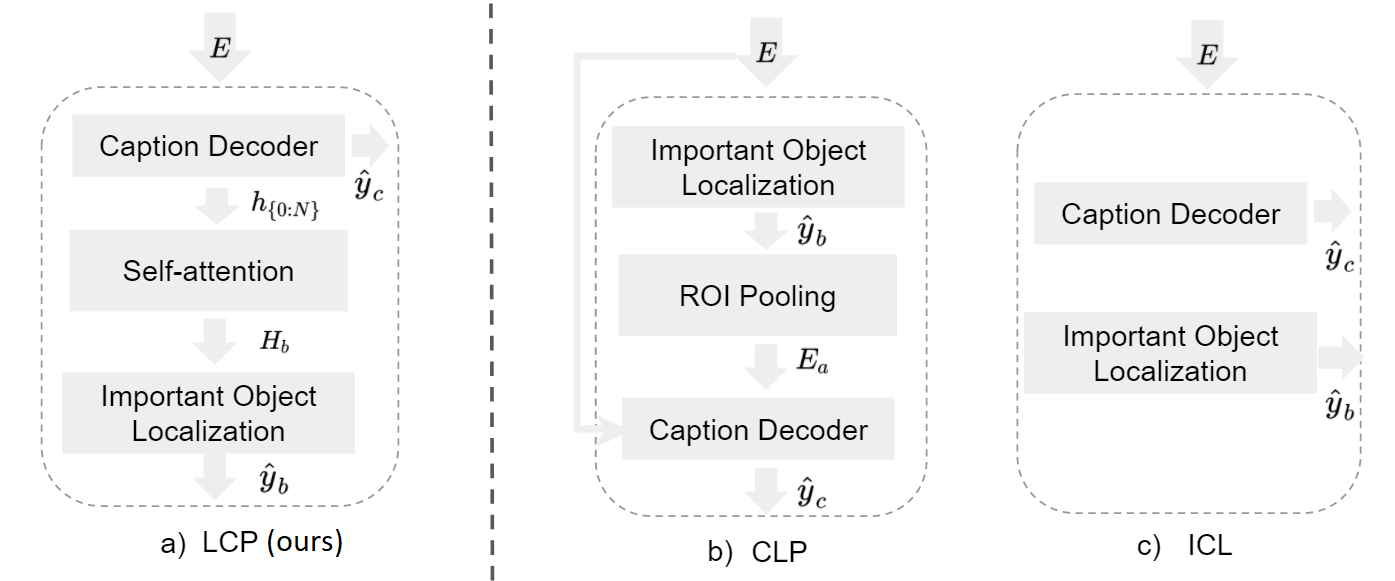}
\caption{Comparison of our decoder (LCP) with that of two baselines (CLP and ICL).}
\label{fig:block_diagram1}
\end{figure}
\subsection{Network Architecture}
\label{sec:net_arch}

The network architecture is shown in Table~\ref{tbl:model_summary}. Both visual encoder and flow encoder use ResNet-101 to extract features using the layers 0-309. Layer 0-308 is taken from original ResNet-101, and we add Layer 309 to generate features with a fixed size of (2048,14,14). In the decoder, Layer 310(a) concatenate both visual and flow features and the output size is (4096,14,14). Layer 310(b) is to convert the word representations from LongTensor to embeddings with a size of 512 with T as maximum sentence length 50 (after padding zeros). Layer 311(a,b) converts the concatenated embedding to initial hidden state (a) and cell state (b) for the LSTM, followed by the cell update in Layer 312, using attention mechanism on the encoded image features, please refer to SAT[30] for more details. Layer 313 (dropout), 314 (fully connected), and 315 (softmax) is the conversion of LSTM hidden state to the output word at each time step. We rollout for T times through Layer 312-315. In practice, to save computation time each sentence is rolled out only till the length of the ground truth sentence during training. In Layer 317, we use the self-attention mechanism explained in the main draft Sec 4 (sub-section decoder), which takes all the hidden states of the LSTM and performs the self-attention operation. The output from self attention is converted to bounding box parameters from Layer 318-319 using fully connected layer (Layer 318) and Sigmoid operation (Layer 319).

\subsection{Block Diagrams}
In Figure~\ref{fig:block_diagram1}, we visualize the block diagram of decoders used for evaluation. We refer to Section 4 and 5 in the main manuscript for overview ours (LCP) as well as other baseline models (CLP and ICL). Details of ROI pooling used in CLP is presented in Fast-RCNN~\cite{girshick2015fast}. 

\subsection{Model Training}
We build our framework using the PyTorch framework with Tesla V100-SXM2-32GB GPUs. The batch size is set to 16, and we use Adam optimizer with a learning rate of 1e-4 for the encoder and 4e-4 for the decoder. The dropout ratio is 0.5 in the word\_dropout\_0 layer in Table~\ref{tbl:model_summary}. The model is trained for 120 epochs, and we report the best checkpoint (\textit{i.e.}, saved using the CIDEr score measured from the validation set) to evaluate the model on the test set. 

\section{Additional Evaluation}

\subsection{Model Robustness}
We ran our best model for three times and computed the standard deviation for the metrics on the test set.
As shown in Table~\ref{tab:std_deviation}, the small value of standard deviation demonstrates that our approach consistently provides robust prediction capabilities.

\subsection{Caption Length vs CIDEr Score}
\begin{figure}[h]
    \centering
    \includegraphics[width=0.45\textwidth]{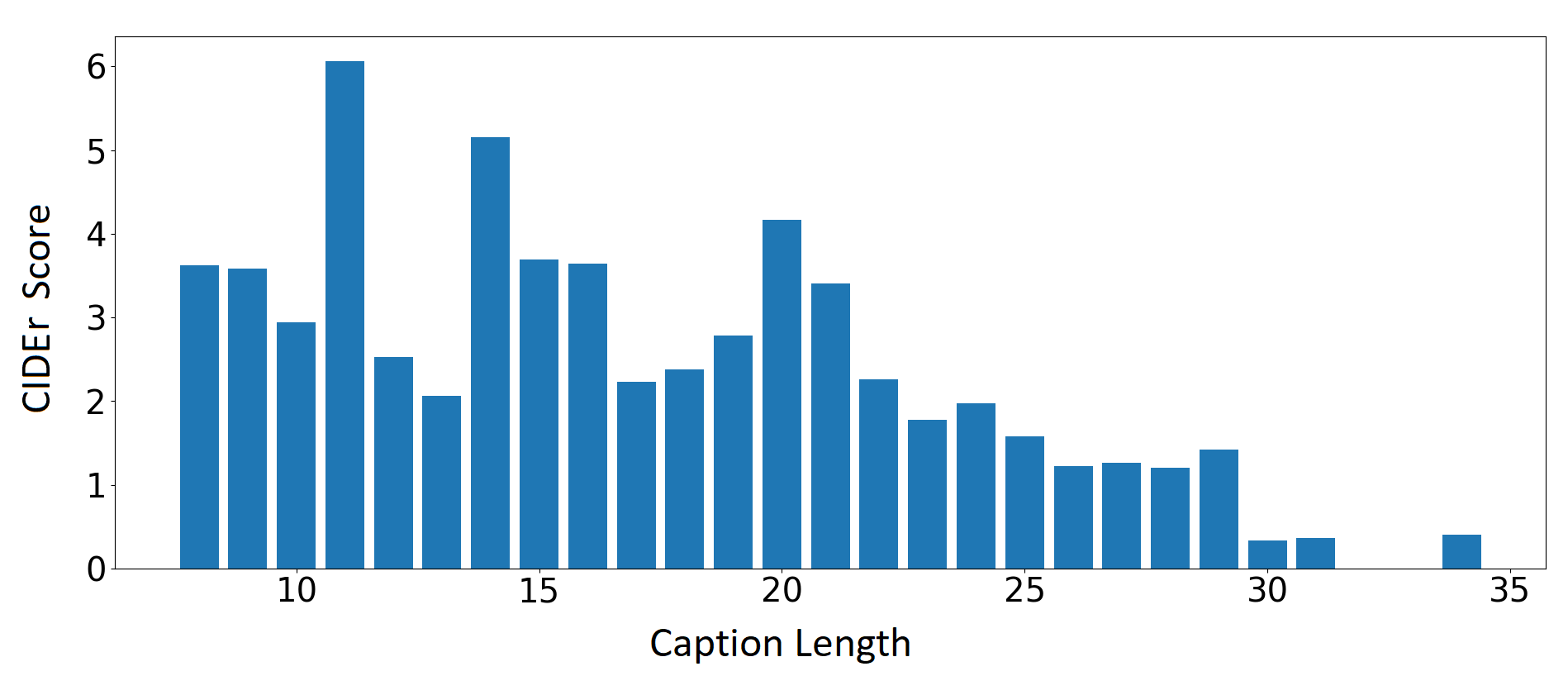}
    \caption{Caption Length vs Average CIDEr Score are evaluated on the test set.}
    \label{fig:len_vs_cider}
\end{figure}
\begin{figure*}[t]
    \centering
    \includegraphics[width=\textwidth]{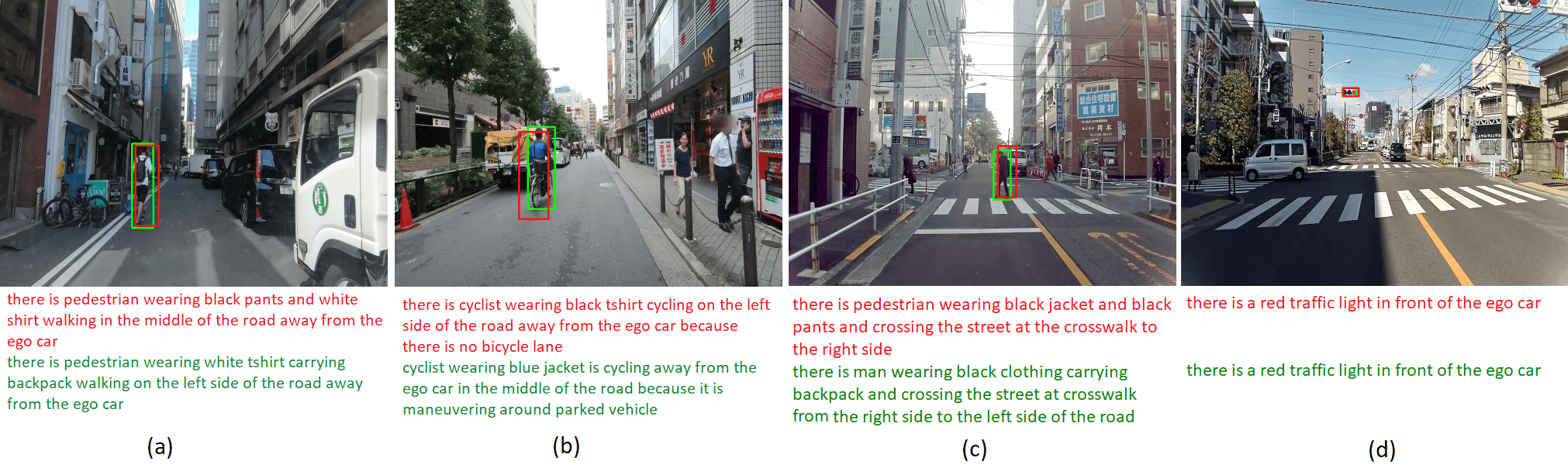}
    \caption{Successful cases. Ground truth in green color and Prediction in red color.}
    \label{fig:qual1}
    \vspace{-0.25cm}
\end{figure*}
In Figure~\ref{fig:len_vs_cider}, we plot the CIDEr score of our model with respect to the length of generated captions using the test set. The CIDEr score is sorted in the order of ascending length. In general the performance degrades with the increase in the length or complexity of the sentence. The increase in performance at several lengths is because the captions are more commonly observed with those lengths.


\subsection{Multi-Task Training}
We compare two metrics, CIDEr and Mean-IOU, while training our model with (LCP Ours) and without (LCP without DTL) multi-task weighting. The captioning metric CIDEr performance is shown in Figure~\ref{fig:val_cider}. The important object identification metric Mean-IOU performance is shown in Figure~\ref{fig:val_iou}.
We weigh the model `LCP without DTL' with $\lambda=15$ in Equation~\ref{eq:loss_sum}, which we found the best performing model from the manual weighting. The model `LCP (Ours)' is trained using the loss in Equation 3. $L_{c}$, $L_{b}$ refers to the caption loss and bounding box loss as mentioned in Section 4 of the main manuscript. As shown in Figure~\ref{fig:val_curves}, `LCP (Ours)' shows the continuous improvement of the performance for both tasks. However, `LCP without DTL' sacrifices the capability of one task (captioning) while improving the other (object identification).
\begin{equation}
    L_{sum}=L_{c}+\lambda*L_{b}
    \label{eq:loss_sum}
\end{equation}


\subsection{Qualitative Analysis}

\begin{figure*}[!h]
    \centering
    \includegraphics[width=\textwidth]{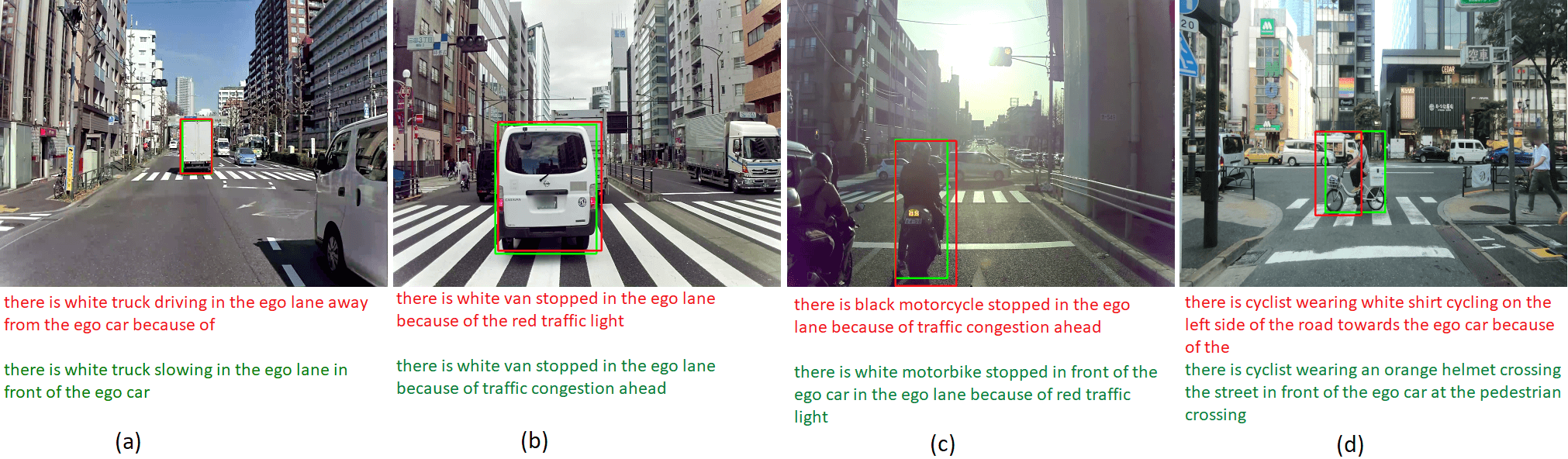}
    \caption{Captioning failures. Ground truth in green color and Prediction in red color.}
    \label{fig:qual2}
    \vspace{-0.25cm}
\end{figure*}
\begin{figure*}[!h]
    \centering
    \includegraphics[width=\textwidth]{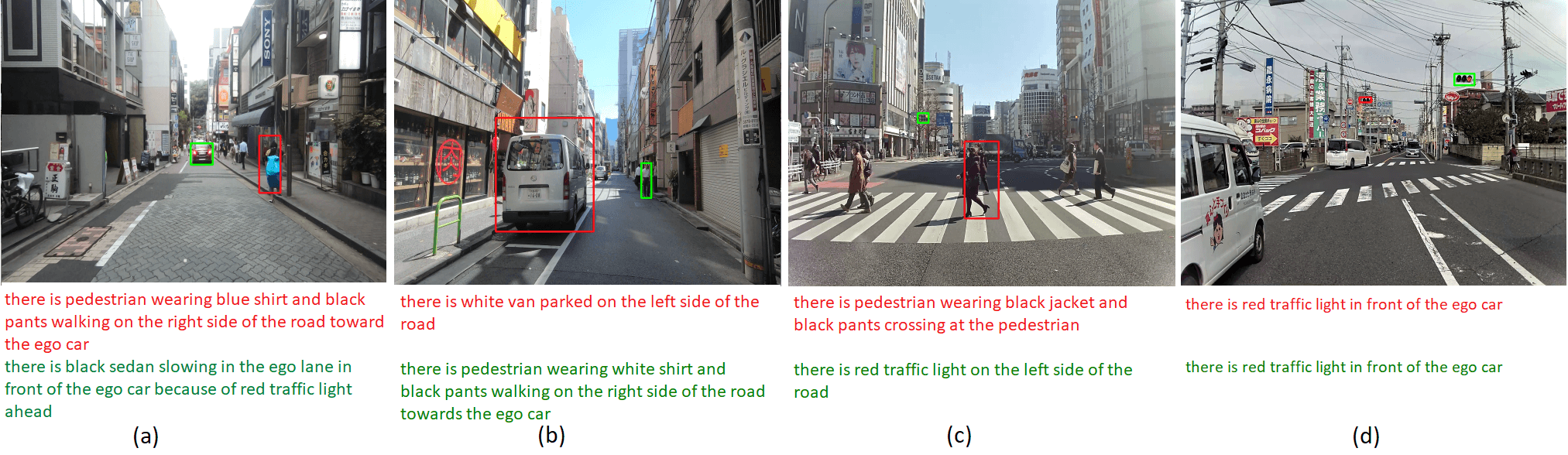}
    \caption{Localization fail cases. Ground truth in green color and Prediction in red color.}
    \label{fig:qual3}
    \vspace{-0.25cm}
\end{figure*}

In Figure~\ref{fig:qual1}, we visualize additional successful cases in addition to Figure 7 shown in the main manuscript.  \ref{fig:qual1}a and \ref{fig:qual1}c are two different pedestrians walking on a narrow road and at the intersection. In both cases, the visual, location, motion direction attributes are generated correctly. Similarly, the cyclist in \ref{fig:qual1}b is identified as an important agent with a good reasoning (`because there is no bicycle lane'). In \ref{fig:qual1}d, the traffic light is predicted correctly even when there is a vehicle in front of the ego car.

In Figure~\ref{fig:qual2}, we show the captioning failures while the localization of the important agent is correct. In \ref{fig:qual2}a, the truck is found correctly but its reasoning is incomplete. In \ref{fig:qual2}b, the network reasoned about stopping as red traffic light. The model might be confused because the traffic congestion is hard to see in this scenario. In contrast, \ref{fig:qual2}c is an example of the opposite case, where our model is hard to see the traffic light as a reason for stopping. In \ref{fig:qual2}d, the cyclist's motion direction is predicted wrong which resulted in incomplete reasoning.

In Figure~\ref{fig:qual3}, localization failures are shown. The network sometimes gets confused to identify the important object when there are multiple objects shown reasonably important in the scene. We show several cases:  `\textit{vehicle$\rightarrow$pedestrian' in \ref{fig:qual3}a}, `\textit{pedestrian$\rightarrow$vehicle' in \ref{fig:qual3}b},  `\textit{infrastructure$\rightarrow$pedestrian' in \ref{fig:qual3}c}, `\textit{infrastructure$\rightarrow$ infrastructure' in \ref{fig:qual3}d }. In scenarios of \ref{fig:qual3}a-c, the identified important object is rather closer to the ego-vehicle whereas the ground truth is annotated very far. The encoder of our model might be hard to extract image features of those agents (as their size is very small). The advancement of the algorithm to cover such cases are our future research direction. In scenario \ref{fig:qual3}d, there exist multiple traffic lights showing the same traffic information. Although our identification is different from the ground truth, the model successfully reasons about the interaction as a caption, which activated braking of the vehicle. 

\section{Limitations}
There has been a significant progress in computer vision and machine learning algorithms for autonomous and cooperative driving. However, explainable models that ground language and vision in this domain is less highlighted. 
In this work, we introduce a new dataset and address risk localization with its reasoning as a language description in driving scenarios where such a model can benefit to situational awareness in autonomous driving and driving assistant systems. 
Our setting of interest and the dataset have an objective of addressing driving risks that implicitly and explicitly influence safety-critical design of intelligent systems. Therefore, real-world applications that adopt such reasoning models may cause injury or death to personnel or loss of property if they fail to make appropriate predictions. Although we provided several failure cases to better understand the limitation of the models and our problem settings, the users should be aware of other failures not explored in this paper. Besides the model, our dataset is labeled by humans, which may incorporate biases due to errors in interpretation. The users are also requested to assess the risk while using our dataset for their safety critical applications.


\end{document}